%% file: main.tex
\documentclass[sigconf]{acmart}

\usepackage[all]{nowidow}
\usepackage{mathrsfs}
\usepackage{boxedminipage}
\usepackage{graphicx}
\usepackage{comment}
\usepackage{enumitem}
\usepackage{multirow}

\usepackage{subcaption} 
\usepackage{soul}
\usepackage{todonotes}

\author{Andrea Coletta}
\orcid{1234-5678-9012}
\author{Matteo Prata}
\author{Michele Conti}
\author{Emanuele Mercanti}
\author{Novella Bartolini}
\affiliation{%
  \institution{Sapienza University of Rome}
  \city{Rome}
  \country{Italy}
}

\author{Aymeric Moulin}
\author{Svitlana Vyetrenko}
\author{Tucker Balch}
\affiliation{%
  \institution{J.P. Morgan AI Research}
  \city{New York}
  \country{USA}}

\begin{document}

\title{Towards Realistic Market Simulations: a Generative Adversarial Networks Approach}

\begin{abstract}

Simulated environments are increasingly used by trading firms and investment banks to evaluate trading strategies before approaching real markets. 
Backtesting, a widely used approach, consists of simulating experimental strategies while replaying  historical market scenarios.
Unfortunately, this approach does not capture the market response to the experimental agents' actions. 
In contrast, multi-agent simulation presents a natural bottom-up approach to emulating agent interaction in financial markets. 
It allows to set up pools of traders with diverse strategies to mimic the financial market trader population, and test the performance of new experimental strategies.
Since individual agent-level historical data is typically proprietary and not available for public use, it is difficult to calibrate multiple market agents to obtain the realism  required for testing trading strategies. To addresses this challenge we propose a synthetic market generator based on Conditional Generative Adversarial Networks (CGANs) trained on real aggregate-level historical data.
A CGAN-based "world" agent can generate meaningful orders in response to an experimental agent.
We integrate our synthetic market generator into ABIDES, an open source simulator of financial markets. 
By means of extensive simulations we show that our proposal outperforms previous work in terms of  stylized facts reflecting market responsiveness and realism. 
\end{abstract}

\maketitle
\renewcommand{\shortauthors}{Coletta et al.}

\section{Introduction}

\noindent 
The standard approach in the evaluation of trading strategies before their adoption in real markets, is to simulate their performance under a wide range of existing historical scenarios --- a technique known as {\em backtesting}.
Despite its common use in trading applications due to its simplicity, the backtesting approach does not allow to simulate market response to experimental agents' actions, which makes it unreliable for  simulating  the execution of trading strategies.

Interactive Agent-Based Simulators (IABS) address the limitations of the backtesting approach by modeling the interplay of individual traders, the agents, of a market system. They are a fundamental tool in financial research for studying and testing trading strategies, while  also providing market behavior explainability  \cite{macal2010tutorial, paulin2018agent}. 
%
An agent-based simulation allows to study the market response to the experimental agents  
and 
enables ``what if'' studies on how their strategy would perform  \cite{balch2019evaluate}. 
IABS are increasingly being used 
for training of 
machine learning based 
trading strategies \cite{spooner2018market, karpe2020multi, buehler2019deep}, and deepen the study of real market phenomena  \cite{borrajo2020simulating, wang2020market, yagi2020analysis}.
While most market simulators are used for internal purposes within trading firms, robust simulators for high-fidelity market simulation environments are becoming available also for the research community. 
One example in this direction is ABIDES (Agent-Based Interactive Discrete Event Simulation) \cite{byrd2020abides}.

Modeling a realistic  market behavior is a major challenge, common to all the described works.
To obtain realistic simulations, these approaches rely on pools of hand-crafted experimental agents to  mimic the complexity of a real market. 
Along this line of research, previous works \cite{byrd2019explaining, gode1993allocative, palit2012can}
analyzed traders' behaviors through market data-sets and described a series of trading strategies.
%
However, a real stock market includes  thousands of participants adopting diverse proprietary trading strategies.
Historical data typically lacks complete details of trader population and of the adopted strategies. Therefore, it is often impossible to exactly  understand
the traders' behaviors, and to   precisely mimic them through simulated agents. 
This   inevitably leads to poorly realistic simulations, and as a result ``what if'' studies may give inaccurate results \cite{vyetrenko2019get}.

To the best of our knowledge, we are first to propose a GAN-based framework that is able to learn an abstract representation of the market that can also react to the observed market state.
The proposed framework leverages a Conditional Generative Adversarial Network (CGAN) \cite{mirza2014conditional, NIPS2014_gan} trained on real historical data, to capture the market’s behavior as a whole arising from the activity of different participants. 
The CGAN is trained to generate  limit orders conditioned by the currently observed market situation.
%
A world agent impersonated by a pre-trained CGAN that is placed in a simulation environment, produces orders to be processed by the exchange. This agent represents the whole ensemble of trading agents of the particular symbol on which it was trained. Such simulation test-bed allows to evaluate any experimental agent, while guaranteeing realism of the produced market trend and market responsiveness to the experimental agent's activity.    
We show that  the market traces generated by the proposed CGAN-based agent preserve the statistical properties of historical data, and at the same time expose a reactive behavior against experimental agents. 

Here we summarize the original contributions of this paper:
\vspace{-0.05in}
 \begin{itemize}
    \item We propose a framework able to automatically learn and reproduce the market behavior, allowing \textit{realistic} and \textit{reactive} simulations. 
    \item We propose a CGAN architecture able to generate synthetic market data, in terms of limit orders, as a function of the features observed from the current market situation, thus guaranteeing both market realism and reactivity to experimental agents. 
     \item We analyze the realism of generated orders and  study the reactivity of the simulation framework to experimental agents by comparing simulations conducted with and without them. 
     \item We compare our CGAN based simulations against a state of the art IABS \cite{vyetrenko2019get}, showing that our architecture outperforms existing approaches, exhibiting more realistic properties in terms of stylized facts of a Limit Order Book (LOB) \cite{cont2001empirical, chakraborti2011econophysics}.
 \end{itemize}
 
\vspace{-0.05in}
\section{RELATED WORK}
IABSs are increasingly gaining attention by the research community, with several works studying simulators and solutions to closely mimic real markets \cite{vyetrenko2019get, byrd2020abides, karpe2020multi, palit2012can, farmer2005predictive}.

Farmer et al. \cite{farmer2005predictive} study market simulations showing how zero intelligence (ZI) agents, which place random orders, are able to reproduce market dynamics (i.e., spread and price) and react to the arrival of new orders. Palit et al. \cite{palit2012can} deepen the study of ZI agents in market simulations, showing that they can also reproduce fat tails and long range dependencies.
\\
In a recent work \cite{vyetrenko2019get}, S. Vyetrenko et al. survey metrics to assess robust and realistic market simulations. The authors study two different agent configurations showing how IABS configurations that still rely on common sense hand-crafted rules may produce unrealistic market scenarios. In the experimental section, we compare our framework against their agent configurations and evaluate the related realism metrics.

Other approaches to achieve realistic simulations focus on automatic learning strategies. An action conditional stochastic forward model (otherwise known as a world model) was previously used to learn market response to trading from historical data \cite{world_models, lidia_world}. The model is implemented via a Convolutional Neural Network (CNN) autoencoder followed by Mixture-Density Recurrent Network (RNN-MDN), and is subsequently used for reinforcement learning trading agent training, but is not tested for simulated realism.

Other works in the literature focus on producing synthetic financial time-series, retaining  statistical properties of historical data, \cite{assefa2020generating, li2020generating, takahashi2019modeling, NEURIPS2019_c9efe5f2, efimov2020using}. 
For example, Junyi Li et al. \cite{li2020generating} propose an approach to generate realistic and high-fidelity stock market order streams, based on GANs. The proposed \textit{Stock-GAN} model employs a Conditional Wasserstein GAN to capture the history dependence of orders. The goal of this study is to generate synthetic data that does  not include features that may link back to the originator of the order in the training data-set, for anonymity reasons.
\\
However, none of these works address the issue of responsiveness and reactivity to experimental agents.

 \begin{figure}
 \centering
   \centering
   \includegraphics[width=0.9\linewidth]{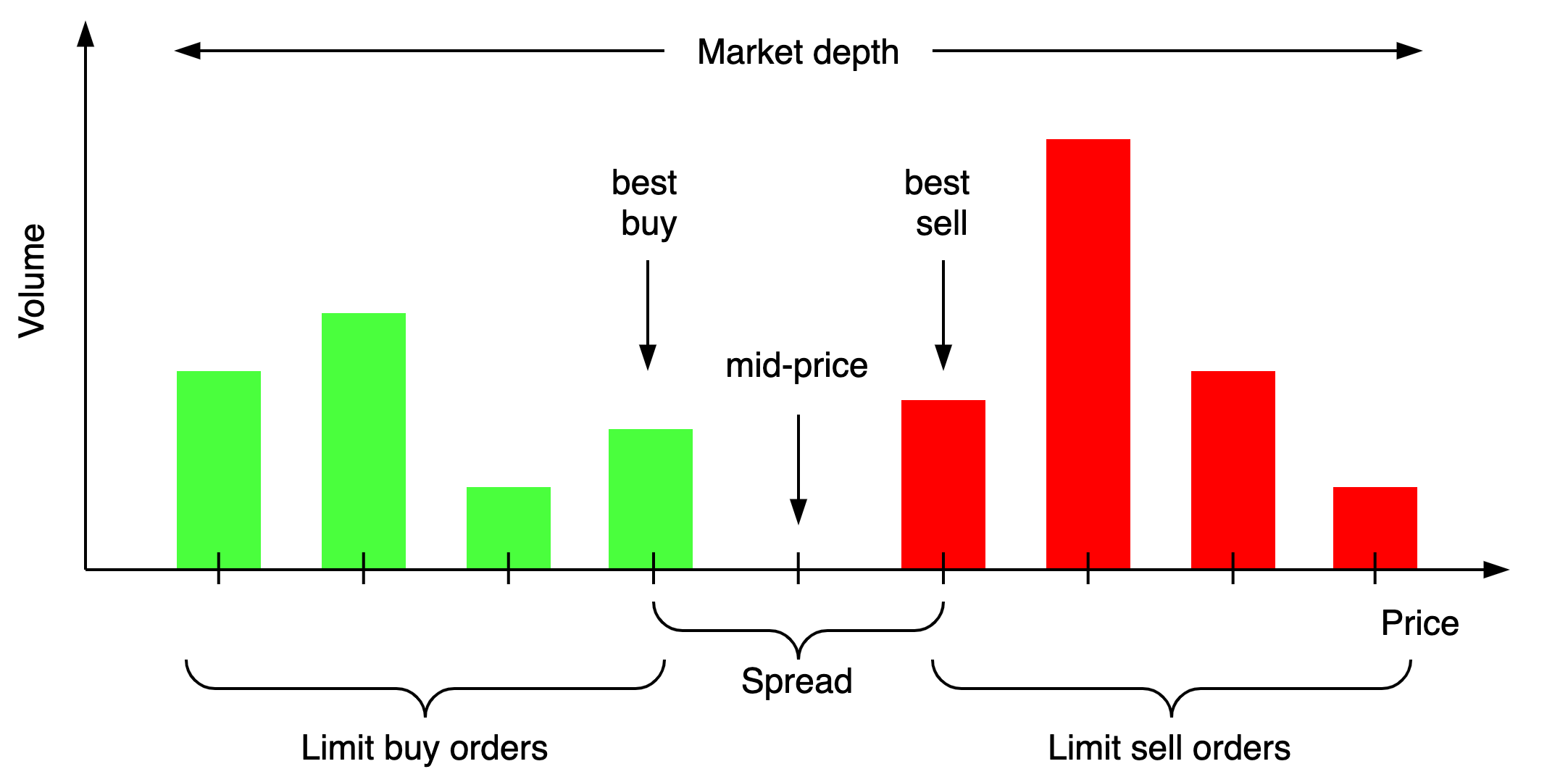}
   \caption{An example of order book.}
 \label{fig:lob}
  \vspace{-0.2in}
 \end{figure}

\begin{figure*}[t]
 \centering
    \begin{minipage}{.49\textwidth}
        \centering
  \includegraphics[width=0.9\linewidth]{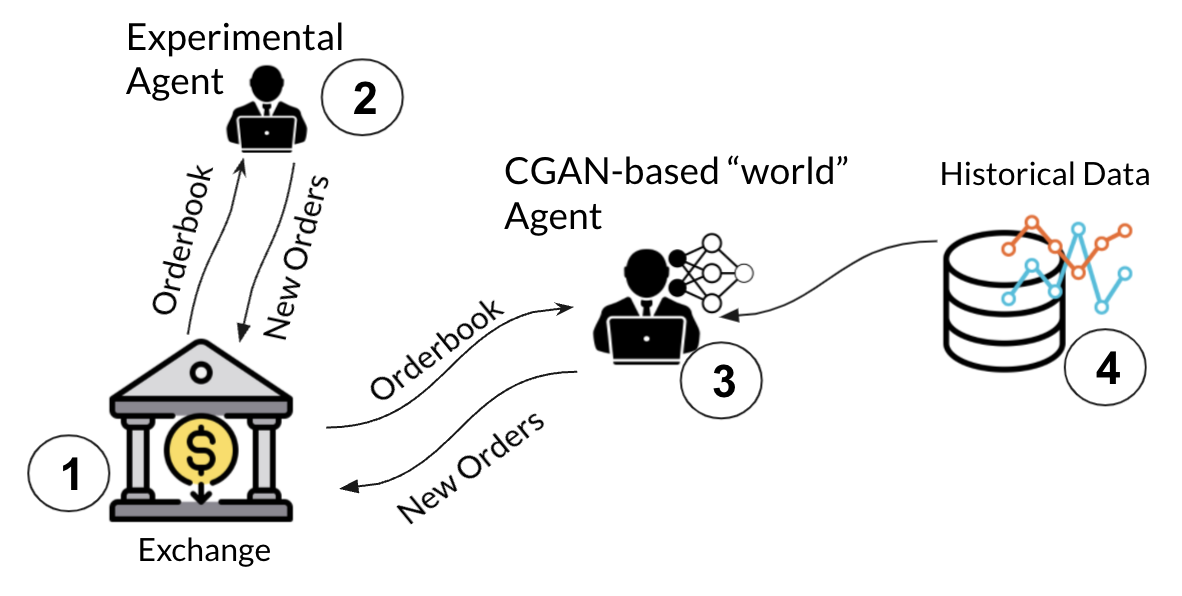}
  \caption{CGAN-Based Simulation Framework. 
  }
  \label{fig:sys_mod}
    \end{minipage}%
    \hfill
    \begin{minipage}{.49\textwidth}
\vspace{-0.15in}
        \centering
  \includegraphics[width=0.78\linewidth]{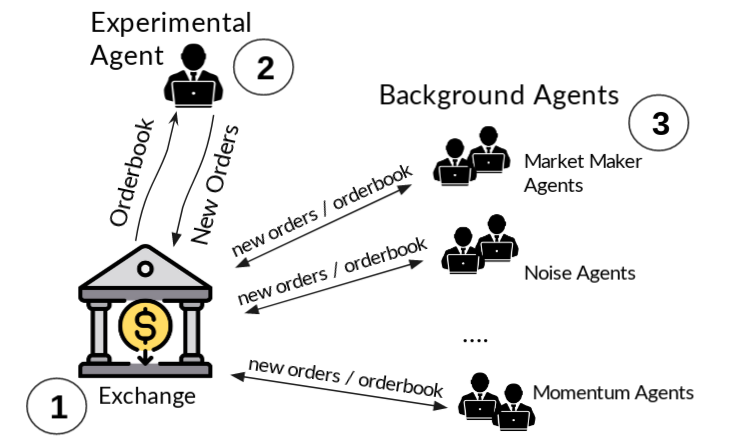}
  \caption{IABS Simulation Framework. 
  }
  \label{fig:iabs_mod}
    \end{minipage}%
    \vspace{-0.1in}
\end{figure*} 

\vspace{-0.05in}
\section{BACKGROUND} 
In this section we briefly review some important pillars that are useful for the reader to approach this paper.
\vspace{-0.05in}
\paragraph{\textbf{Orders and Limit Order Book}} 
In  financial markets, traders can participate in the auction mechanism by placing orders to \textit{buy} [bid] or \textit{sell} [ask] shares of the financial assets of choice. There are two major types of orders: (1) \textit{market orders}, which are executed immediately at the best available price, i.e. highest buy (lowest sell) price in case of a sell (buy) market order; (2) \textit{limit orders}, which, unlike the market orders, include the specification of a desired target price. A sell (buy) limit order will be executed only when is matched to a buy (sell) order whose price is greater than (lower than) or equal to the target price. The exchange matching engine  is responsible for matching the incoming buy and sell orders issued by the trading agents.
Figure \ref{fig:lob} shows the LOB of a financial asset, which is an electronic record of the orders issued by the traders. LOBs are maintained at financial exchanges, such as NASDAQ and NYSE, and act as an indicator of the supply and demand of a particular asset in time. They are organized by price levels and  are continuously updated as new orders are placed. 
\vspace{-0.05in}
\paragraph{\textbf{Stylized Facts of financial markets}}\label{sec:stylized}
From the analysis of the market and LOB evolution over  time, several metrics (e.g., spread, mid-price, relative return) and other statistical properties (e.g., absence of autocorrelation or fat-tailed distribution of asset relative returns) can be derived.
These statistical properties capture the market behavior over different time periods, and are often referred to as \textit{stylized facts} \cite{cont2001empirical, chakraborti2011econophysics}.

One can calculate stylized facts for the time-series produced by agent interactions for real and simulated markets, and use them as metrics of simulation realism \cite{vyetrenko2019get}. 
In our work, we consider a set of the most commonly used stylized facts to evaluate how our CGAN-based "world" agent is able to reproduce characteristic features of a real market. 
We study the following stylized facts:
\begin{itemize}
    \item \textbf{Absence of Autocorrelation}. The linear autocorrelation of the return $C(\tau) = \text{corr}(r(t, \Delta t), r(t + \tau, \Delta t))$ rapidly decays to zero in few minutes and becomes insignificant for periods longer than 20 minutes.
    \item \textbf{Volatility Clustering}. High-volatility events tend to cluster in time, showing a positive autocorrelation over several days of trading.
    \item \textbf{Volume/Volatility correlation}. The trading volume is positively correlated with volatility.
    \item \textbf{Heavy tails and aggregation normality}. The asset returns follow a normal distribution with fat tails. Notice that, fat tails disappear when the time period $\Delta t$ use to compute the returns increases.
\end{itemize}

 \vspace{-0.1in}

\paragraph{\textbf{Generative Adversarial Networks (GANs)}}

GANs \cite{NIPS2014_gan} are a powerful subclass of generative models and were successfully applied to image generation and editing, semi-supervised learning, and domain adaptation. 

With GANs, we have a two-player game between the generator $G$, learning how to generate samples resembling real data, and a discriminator (or critic) $D$, learning to discriminate between real and generated data. Both players aim to minimize their own cost, and the solution to the game is the Nash equilibrium where neither player can improve its cost unilaterally \cite{NEURIPS2018_e46de7e1}.
In this paper, we consider a generator $G$ to learn how to produce new orders, matching real orders statistical properties. 
In the GAN framework the model learns a deterministic transformation $G$ of a simple distribution $p_z$, with the goal of matching the real data distribution $p_d$.
On the other hand, $p_z$ is the distribution of the generated data, that is implicitly defined as $\tilde{x} = G(z)$, where $z \sim \mathcal{N}(0,\,1)$ is the input noise to the generator, sampled from a normal distribution. 
Random noise as input to $G$ ensures variance of the generated data. 
The transformation $D$ is trained to maximize the probability of assigning the correct label to both training examples $x \sim p_d$ and $\tilde{x} \sim p_z$, while at the same time $G$ is trained to minimize the probability of $D$ assigning the correct label. 
The  novelty of our work is the design of a synthetic market order generator whose behavior reacts to the activity of experimental agents, resembling  realistic markets. 
%
To this end, while in its most basic form a GAN generator takes random noise as its input to transform noise into a meaningful output, we consider a Conditional Generative Adversarial Network (CGAN) \cite{arjovsky2017wasserstein} 
which takes as input also several features that characterize the emulated market behavior.

This translates into a  joint two-player minimax game that can be described by the following dynamic:
\vspace{-0.05in}
\begin{equation}
\label{eq:loss}
   \min_G \max_D \, \underset{x \sim p_d}{\mathbb{E}}[\log(D(x|y))] - \underset{z \sim \mathcal{N}(0,\,1)}{\mathbb{E}}[\log(D(G(z|y)))], 
\end{equation}
where $y$ is some extra information that conditions the players' behavior. The minimax game derives from the cross-entropy between the real and generated distributions. Thus $\log(D(x|y))$ and $\log(D(G(z|y)))$ reflect the likelihood  that the discriminator correctly classifies real data $x$ given $y$, or generated data $G(z|y)$ respectively. The generator $G$ cannot directly affect the $\log(D(x|y))$ term, thus for him minimizing the loss is equivalent to maximizing $\log(D(G(z|y)))$ as it is a negative additive quantity. 
\section{CGAN-Based Simulation Framework}
The proposed simulation framework, represented in Figure \ref{fig:sys_mod}, includes four main components:
\vspace{-0.05in}
\begin{enumerate}
    \item The electronic market \textbf{exchange}, which  handles incoming orders.
    \item One or more \textbf{experimental agents}, which operate the trading strategy under investigation.
    \item The \textbf{CGAN-based "world" agent}, which is responsible for emulating the market behavior, by generating new orders as a function of the current market state.
    \item The \textbf{historical data}, which is used to train the CGAN agent and to initialize the simulation. 
\end{enumerate}

In contrast to IABS simulations where experimental agents interact with multiple diverse market participants via an exchange to mimic a real market (Figure \ref{fig:iabs_mod}), we consider a unique CGAN-based agent, acting as the whole trader population (Figure \ref{fig:sys_mod}). 
In fact, while market behavior is typically modeled through multiple heterogeneous agents in IABS simulations, these approaches fail in guaranteeing the necessary realism and responsiveness \cite{vyetrenko2019get}.

The lack of knowledge about traders  and their strategy in historical data, motivates poorly realistic simulations. 

In our simulation framework, a CGAN-based agent is trained on historical data to emulate the behavior resulting from the whole set of traders.
It analyzes the order book entries and mimics the market behavior by producing new limit orders depending on the current market state. 
Thus, the experimental agents interact with the CGAN-based agent through the exchange, enabling the test and the study of their strategy against a responsive and realistic market.  

\vspace{-0.05in}
\subsection{Electronic Market Exchange Simulator} 
The electronic market exchange handles the LOB of a set of securities, applying a price-then-FIFO matching rule for the trading orders issued by the agents.
For our simulations we use ABIDES \cite{byrd2020abides}. It is an open-source simulator that provides a selection of background agent types (such as market markers, fundamental value investors, momentum traders, etc.), a NASDAQ-like exchange agent which lists any number of securities for trade using a LOB with price-then-FIFO matching rules, and a simulation kernel which manages the advancement of time and handles all inter-agent communications.

\vspace{-0.05in}
\subsection{Experimental Agents}
An experimental agent is an entity of a financial IABS, representing a financial trader adopting a specific trading strategy. It owns a portfolio, pending orders and initial cash.
Experimental agents are implemented to evaluate the performance of their  trading strategy, taking into account agent actions' impact in the estimated profitability, or to pursue ``what if'' studies.
\\
Any trading strategy can be implemented inside an experimental agent introduced in the simulation, including machine learning based ones \cite{karpe2020multi}. 

For instance, in our  experimental analysis in Section \ref{sec:exp}, we use an experimental agent which 
injects orders so as to reach a given cumulative volume within a predefined interval of time. We refer to this agent with the name of Percentage of Volume (POV) experimental agent \cite{vyetrenko2019get}.
We use this agent in our simulations  to  evaluate the responsiveness of our CGAN-based "world" agent. 

\vspace{-0.05in}
\subsection{CGAN-based World Agent}

The core component of our framework is the CGAN-based "world" agent, which mimics the market behavior. To this end, our CGAN architecture takes as input a set of features, representing the current market condition, and generates the next order to feed into the exchange. The new order is processed by the exchange and the simulation advances, generating a new market state, and a subsequent new order from the CGAN. 
Thus, while common GAN models output data from input random noise, the proposed CGAN, in addition, is fed with some meaningful features to condition the data generation. 
We consider a similar architecture of Junyi et al. \cite{li2020generating}, based on the conditional WGAN \cite{mirza2014conditional, arjovsky2017wasserstein}, adapted to cope with our task and to react in a simulated environment. 

\paragraph{\textbf{Output data}} 
The output of the proposed CGAN-based "world" agent is what CGAN generator $G$ generates, that is a single order $\tilde{x}$. Formally the output is a 4-tuple $\tilde{x} =$ \textit{(price, volume, direction, time)}, where: \textit{direction} $ \in \{-1, 1\}$ represents sell and buy orders respectively; 
and \textit{time} is the interarrival time between two consecutive orders, more precisely it is the offset w.r.t. the previous order. 
After this \textit{time}, the order is placed into the exchange, and the CGAN is triggered to generate another order.

\paragraph{\textbf{Input data}}\label{sec:normalization}
The input of the proposed CGAN-based experimental agent is made of a vector $y$ responsible for the conditional nature of the method and random noise vector $z$ to provide diversity to generated traces. 

In more details, a vector $y$ is used to condition the generator output. At time $t$ the $i$-th element of the vector with $i \in [1, N]$, represents a series of 10 features linked to a trading order issued at time $t-i$. Specifically, $y \in \mathbb{R}^{10\cdot N}$.
The features are the 4-tuple mentioned in the previous paragraph representing the order details, that is \textit{(price, volume, direction, time)}, concatenated to the 6-tuple made of the following features \textit{(\textit{best-bid price}, \textit{best-bid volume}, \textit{best-ask price}, \textit{best-ask volume}, \textit{mid-price}, \textit{time-period})}. 
As shown in \cite{li2020generating}, most of the sent orders are strongly driven by the current mid-price and order book state, making these features particularly adequate to influence future orders. 
In order to avoid instabilities in the training process of the CGAN, we normalize $y$ with a min-max scaler, between -1 and 1. As for the \textit{size}, \textit{time}, \textit{best-bid volume}, and \textit{best-ask volume} they are pre-processed with a box-cox transformation \cite{sakia1992box} in order to reduce anomalies and normally distribute the data.

 \begin{figure}[t]
 \centering
   \centering
   \includegraphics[width=1\linewidth]{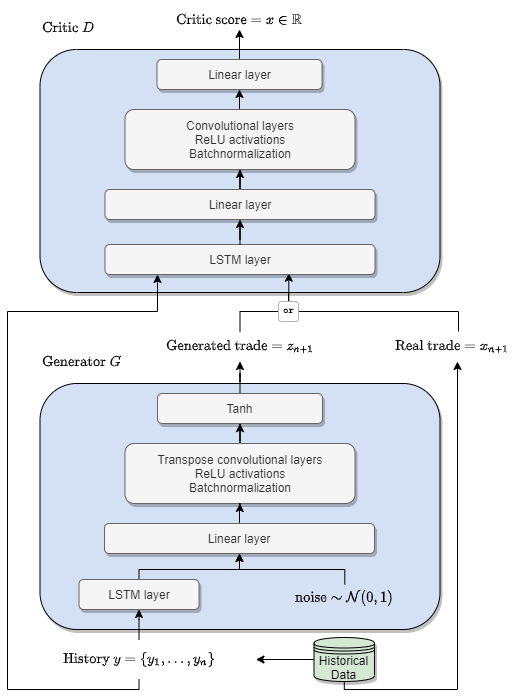}
   \caption{CGAN Architecture.}
 \label{fig:cgan_arch}
 \vspace{-0.2in}
 \end{figure}
Finally, features vector $z$ constitutes part of the input. This vector is made of $N_f$ noise features sampled from a normal distribution. Notice that the concatenation between noise and historical information of the market is necessary. The latter maintain fidelity of the generated data with respect to the historical data: our agent has to issue seemingly real orders, and this can be done only if it has up-to-date knowledge of the market state. Instead, the noise guarantees significant variance in the generated new orders.
%
Noise samples vary at varying simulation seeds, allowing the generation of diverse market behaviors. 
In the experiments, we considered $N = 50$ and $N_f=50$.

\paragraph{\textbf{CGAN Architecture}}

We want our model to generate realistic orders providing responsiveness conditional on the current market state. To do so, we use a CGAN \cite{mirza2014conditional}. 
%
%
Input noise from the prior and extra information $y$ (i.e., features about current market situation) are combined and fed into both the generator and the discriminator. 
%
State of the art Wasserstein Generative Adversarial Network with Gradient Penalty (WGAN-GP) is adopted. 
In order to address training instabilities due to exploding and vanishing gradients, we introduce additional terms and constraints in the loss function as suggested in \cite{gulrajani2017improved}.
In fact, loss function is constructed using the Kantorovich-Rubinstein duality \cite{villani2008optimal}, and considering as discriminator a 1-Lipschitz function (i.e. a function which norm of the gradients is upper-bounded by 1).
Thus the used loss is:
\begin{multline}
\label{eq:gp}
\underset{G}{\min}\,\underset{D}{\max}\, \underset{x \sim p_d}{\mathbb{E}}[D(x|y)] - \underset{z \sim \mathcal{N}(0,\,1)}{\mathbb{E}}[D(G(z|y))] + \\ \underset{z \sim \mathcal{N}(0,\,1)}{\lambda \mathbb{E}}[(\Vert \nabla D(G(z|y)) \Vert _2 -1 )^2]
\end{multline}
where the weight $\lambda$ is set to 10 as in \cite{gulrajani2017improved}.

Figure \ref{fig:cgan_arch} shows the architecture of the CGAN. The generator $G$ is responsible for producing a trading order $\tilde{x}_{n+1} \sim G(z|y)$ which at best resembles characteristics of a real order. The input of the generator is, the normally distributed noise $z \sim \mathcal{N}(0,\,1)$ and the historical data vector $y$ introduced in the previous section. 
Before going through the generator, historical data vector $y$ is fed into a Long Short-Term Memory (LSTM) \cite{hochreiter1997long} to capture the long-term dependencies among historical traces features and producing a compact encoding. The output of the LSTM is concatenated with the normally distributed noise and passed through a series of convolutional layers. CNNs are highly popular due to their ability to detect patterns that are invariant for some transformation, depending on the used filter. We believe that these layers allow to extract relevant information from the LSTM output, which has condensed the history and features of the current market scenario.
The discriminator $D$ takes as input data coming either from the generator or from the data-set made of real orders. It outputs a real number representing the distance between the input order and a real order.

\vspace{-0.05in}
\subsection{Historical Data}

The role of historical data in our proposed simulation framework is twofold. First, our CGAN model needs order book data and order messages for the training. The training is carried out in an unsupervised fashion; the CGAN extracts sequences of historical orders, trying to generate a plausible next order.
The training can be either performed on a single stock during a very tight time window (e.g., one/two days), or on multiple stocks and periods (e.g., one month). In the former way, the trained network can be leveraged to reproduce very specific trends of a certain stock, while in the latter the trained CGAN can be used to emulate more general market trends.

In case of a single stock testing, the historical data is also used to initialize the simulation. If we want to simulate a particular symbol and day, we want to condition the CGAN-based agent with real data at the beginning of the trading day. The input data let the framework evaluate the market features, and feed them in the CGAN to generate coherent orders for that day. The historical data is only needed to start the simulation, as the CGAN-based agent needs the input market condition (i.e., features $\mathbf{y}$) to generate new orders.  
However, if we want to simulate a scenario that is not linked to any symbol nor day, it suffices to launch a run with experimental agents that have the most diverse logic, to produce preliminary data in the exchange, which are then fed into the CGAN to start the generative process.

Notice that, considering the input features, an order book with at least 1-level is needed, along with all the orders submitted to the exchange. However, as our goal is to mimic the collective traders' behavior (not per trader), no sensitive market data are required. 

In our experiments we use data from LOBSTER \cite{lobster}, an online limit order book data tool that provides high-quality limit order book data, which are reconstructed from NASDAQ traded stocks.  It is worth noting that LOBSTER data has a specific format that doesn't provide the full picture. We use a 20-level data-set. It is constituted of: (1) the order book snapshot history (price and quantity for each level) for the first 20 levels on each side (there is a new snapshot every time a price of quantity changes in these levels); (2) the stream of messages (add order, cancellation etc) related to the levels displayed in the order book. 
Liquidity coming from orders placed beyond 20th level is accounted for in the order book when that price level becomes one of the 20 first, but the message that created it is not represented in the data-set. It is a necessary approximation for us in order to make computations tractable.

\vspace{-0.05in}
\section{Experiments}\label{sec:exp}
In this section we evaluate our framework under three main aspects. 
First, we study our CGAN model performance in terms of training convergence and statistical similarity of the generated orders to historical ones.
Then, we evaluate the main novelty of our framework which is the ability to react to an experimental agent. In particular, we show how the CGAN-based "world" agent reacts to the burst of orders issued by a POV experimental agent. 
Finally, we evaluate the realism of our CGAN-based simulations against a state of the art approach. In particular we consider the work of S. Vyetrenko et al. \cite{vyetrenko2019get} which proposes pools of hand-crafted experimental agents to  mimic the complexity of a real market, and studies statistical properties of the resulting synthetic traces.

\input{convergence}
We implement our framework extending ABIDES \cite{byrd2020abides}, an open-source multi-agent simulator. 
We trained the model on order book entries from \textit{AAPL} and \textit{TSLA} on 2 trading days. Each training process lasts 4 days, running for 200 epochs on a NVIDIA V100 GPU \cite{nvidiav100}. The datasets are obtained from the LOBSTER archive \cite{lobster}. 
Being a conditional model, for the first minutes of simulation, the CGAN "world" agent is fed with historical data coming from the training set alone.

\vspace{-0.025in}
\subsection{CGAN Model Performance}
To evaluate the performance of our CGAN model we compare the generated synthetic orders against the real training data, by looking at their statistical properties. 
In particular, we consider the order distributions over the \textit{price}, \textit{volume}, \textit{direction} and \textit{inter-arrival time}. 
In fact, in order to assess the goodness of a generative model, it is usual to compare the statistics of real and synthetic data, as the loss value alone is not a good indicator of convergence \cite{gulrajani2017improved}.

\begin{figure}[b]
 \centering
   \centering
  \vspace{-0.2in}
   \includegraphics[width=0.75\linewidth]{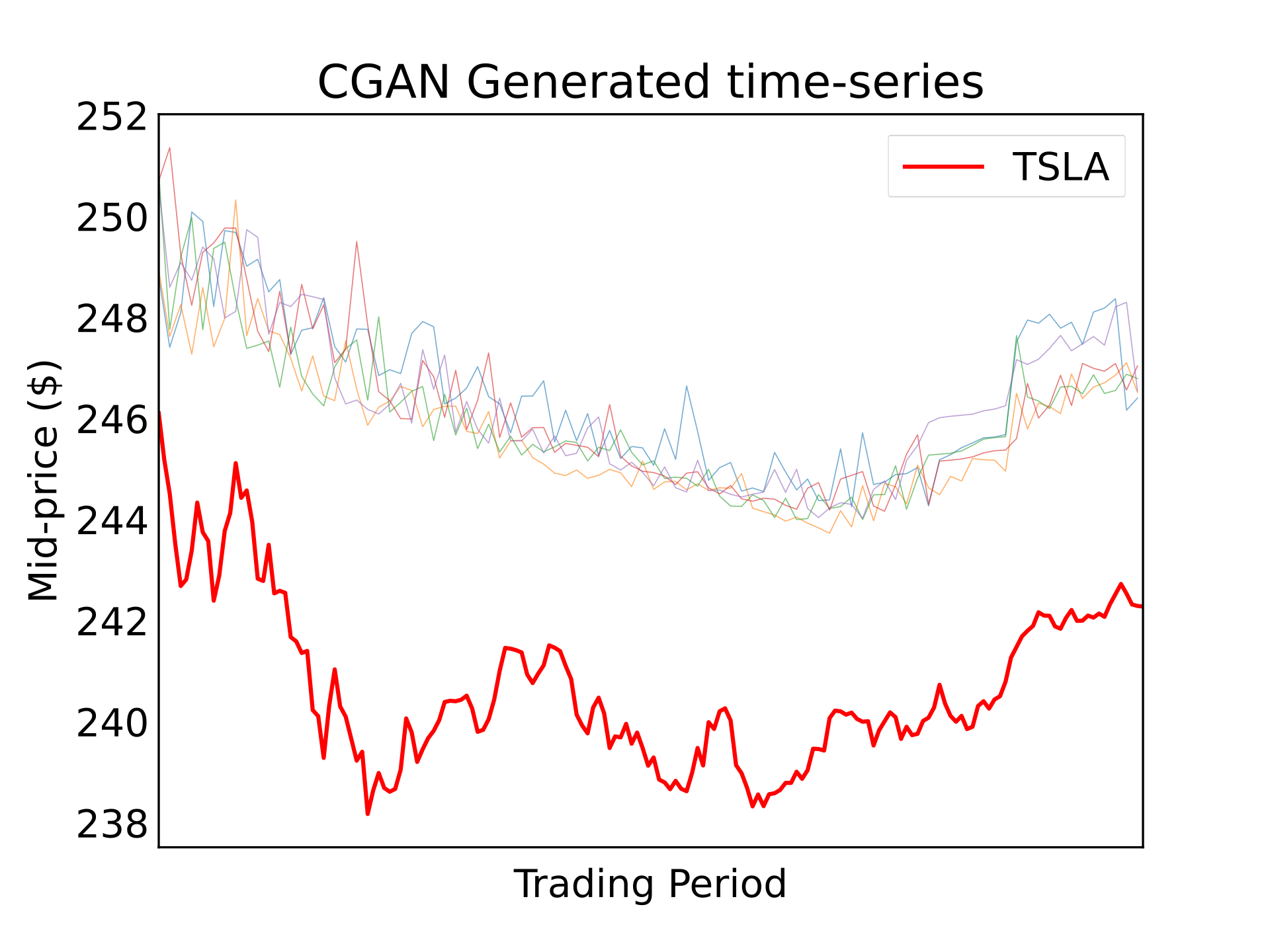}
   \caption{Comparison of multiple CGAN synthetic traces (thin colored lines) and real TSLA real trace (fat red line).}
 \label{fig:tsla_generated}
  \vspace{-0.2in}
 \end{figure}
 
Figure \ref{fig:conv_study} shows the distributions of several useful features for the synthetic and real traces. Specifically,
the columns from left to right, plot the \textit{volume} (i.e., number of shares), the \textit{price}, the \textit{direction} (i.e., buy or sell order), and the \textit{interarrival time} distribution of the orders.
The real data refers to TSLA stock on the 2nd and 3rd of May 2019. All the order features are normalized as in Section \ref{sec:normalization}. The x-axis shows both the normalized form between $[-1, 1]$  and the unnormalized value. 

The first row shows the distributions at the beginning of the training process (epoch number 3), in which the CGAN is clearly unable to generate realistic orders, mostly producing random output.
The second row shows an intermediate advancement of the training process (epoch number 50). The CGAN learns to mimic the distributions more accurately, especially for direction and inter-arrival time. 
Finally, the third row shows a much more stable CGAN agent, achieved after 120 training epochs. The agent generates orders resembling real ones: direction, inter-arrival, price distributions show a pattern that is close to real. Orders volumes are discretionary for the traders, thus it is intrinsically hard to match the real distribution. 

Figure \ref{fig:tsla_generated} shows the CGAN "world" agent ability to replicate trends observed during training, by generating orders that lead to a price pattern that is close to that of the training data. The CGAN was trained on a single stock with a tight time window (i.e., two days).
The generated trends allow pursuing studies on the specific historical scenarios.
The chart shows the training data (solid red line) and 5 generated traces (colored lines), which are conditioned, at the beginning of the generation process with the historical data.
\vspace{-0.05in}
\subsection{Market Impact Experiment}
We demonstrated that our CGAN model is able to generate orders and time-series with properties resembling those of real historical traces. An additional novelty of our framework is the ability to represent all the market participants as a whole and respond to the strategy of an experimental agent, thus allowing interactive simulations. 
In particular, our CGAN-based "world" agent is trained to generate trading limit orders conditioned by the currently observed market situation. Therefore, as the orders submitted by any experimental agent alter the market and the order book, they also condition the CGAN agent and the evolution of the simulation (as shown in Figure \ref{fig:sys_mod}).

In order to evaluate the reactivity of our framework to the activity of an experimental agent, we craft an experiment in which a POV agent submits a burst of \textit{buy} orders \cite{vyetrenko2019get, balch2019evaluate}. 
A $\lambda$-POV strategy is defined by a percentage level $\lambda \in (0, 1]$, a wake up period $\Delta t$, a direction, and a target quantity of shares $\tau$. The agent wakes up every $\Delta t$ time units, observes the transacted volume in the market since its last wake up $V_t$, and issues buy or sell orders of a number of shares equal to $\lambda V_t$, it does so until $\tau$ shares are transacted. 
In our experiments we consider three different $\lambda$-POV agent configurations, with $\lambda \in [$0.01, 0.1, 0.25$]$, and a wake up period $\Delta t = 1$ min.
We allow the agent to place only buy orders between 10:30 and 11:00, and we set $\tau$ equal to the mean daily volume. Thus, the POV agent will trade around 30 times, and it will stop after 11:00 or when it reaches $\tau$ transacted shares. 

 \begin{figure}[t]
 \centering
   \centering
   \includegraphics[width=0.75\linewidth]{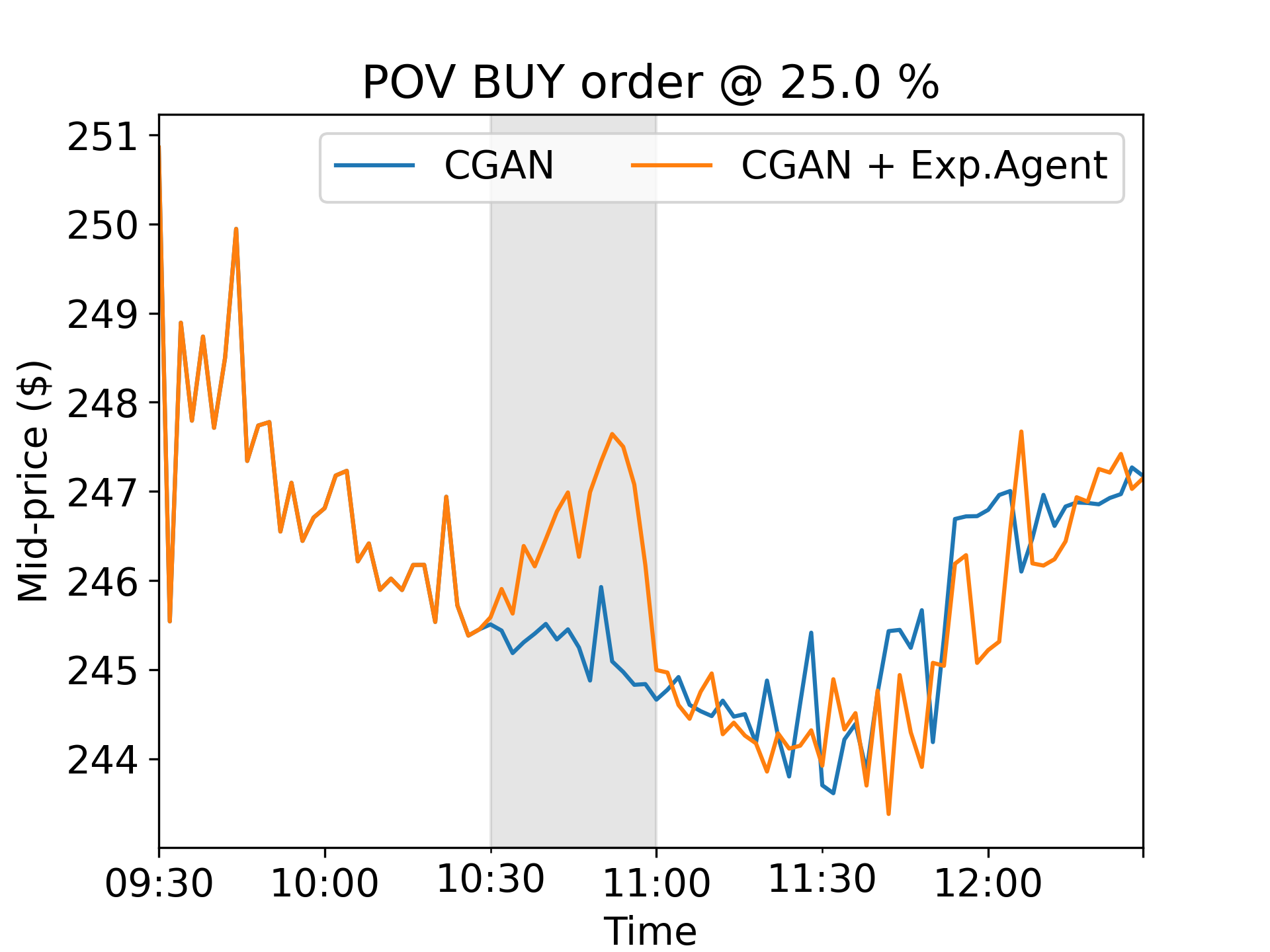}
   \caption{Comparison of a generated time-series with (orange) and without (blue) a $0.25$-POV execution agent, placing order from 10:30 AM to 11:00 AM.}
 \label{fig:impact_example}
 \vspace{-0.1in}
 \end{figure}
  
Figure \ref{fig:impact_plot} shows the impact of POV orders on the stock price, in our CGAN-based framework (right) and in a classic market replay (left) \cite{balch2019evaluate} at varying $\lambda$. 
The plots show the normalized mid-price difference, between the simulations with and without POV agent. The thick blue line represents the median mid-price difference, while the blue shaded regions indicate the 50\% and 90\% quantiles. Results are averaged over 50 runs. 
It is interesting to notice the substantial deviation of the mid-price during and subsequent to the POV agent activity (from 10:30) with the CGAN-based approach w.r.t. the market replay. 
In the market replay scenario, the small deviations are due to new orders in the historical order book being replayed, and rapidly vanish. 
We conclude that our framework can adapt and react to new orders, showing new unobserved behavior. We believe that this is the first step towards reactive IABS \cite{balch2019evaluate}, without the explicit need of hand-crafted pools of experimental agents.

Figure \ref{fig:impact_example} shows the mid-price traces resulting from a single market impact experiment.
%
%
The orange and blue lines are the synthetic data generated by our framework with and without a $0.25$-POV agent, respectively. It is interesting to notice that in the scenario with the POV agent, the mid-price exhibits a substantial deviation w.r.t. the mid-price in the scenario without the POV agent, also showing a common mean reversion trend to its average levels.

\input{impact}
  
\subsection{Realism through Stylized Facts}
Finally, to assess the realism of generated simulations, we train and compare our framework against TSLA historical data, and two state of the art IABS configurations proposed in the work \cite{vyetrenko2019get}. The first IABS configuration includes 5000 noise and 100 value agents; whereas the second configuration comprises 5000 noise, 100 value, 1 market maker, and 25 momentum agents. Both the configurations use historical data to compute the fundamental value of the traded stock.
\footnote{We refer to the work in \cite{vyetrenko2019get} for further details on the configurations and description of the agent strategies.}

Figure \ref{fig:styl_facts} shows the comparison between the three simulation approaches and historical data, in terms of stylized facts (see section \ref{sec:stylized}).
The upper left chart shows the one-minute log-returns distributions. Almost all the simulations show the \textit{aggregation normality} property, but CGAN-based framework (green) is the most similar to historical data.  
The upper right chart shows the \textit{volatility clustering}, as a function of time lag. Also here the CGAN-based framework (green) is the most similar to historical data, showing a slight decay in the correlation as time lag increases. 
The \textit{Volume/volatility correlation} is instead shown in the bottom left chart, in which both the simulations fall short to capture the historical trend. However, the CGAN framework shows the most promising performance, with a similar trend to the historical data, but shifted to the left.
Finally, the bottom right chart shows the \textit{absence of auto-correlation} in which almost all the simulations capture the historical trend, but our framework is still the closest to the real data.

In summary, our CGAN-based framework outperforms the state-of-art configurations, with more realistic stylized facts, close to historical data.

\input{styl_facts}

\color{black}
\section{Conclusion}
In this paper, we provided a novel  framework for synthetic market order generation in agent-based simulations. The purpose of our framework is to provide realistic market scenarios to be used for testing trading strategies, while still ensuring a reactive behavior to the activity of the experimental agents.
Our framework learns the market behavior from historical data-sets, and models all the market participants as a whole, that is a single agent reproducing the entire market behavior and allowing realistic and reactive simulations. 
In particular, we propose a CGAN to generate synthetic market orders as a function of the current market state. Integrated in a simulator, our CGAN-based "world" agent can generate meaningful orders in response to an experimental agent. 
By means of extensive simulations we show that our framework outperforms previous IABS approaches based on hand-crafted market configurations, showing better market realism. 
We also show that our approach ensures a responsive behavior of the simulated market to the activity of experimental agents. 

\section*{Disclaimer}
This paper was prepared for informational purposes by the Artificial Intelligence Research group of JPMorgan Chase \& Co. and its affiliates (``JP Morgan''), and is not a product of the Research Department of JP Morgan. JP Morgan makes no representation and warranty whatsoever and disclaims all liability, for the completeness, accuracy or reliability of the information contained herein. This document is not intended as investment research or investment advice, or a recommendation, offer or solicitation for the purchase or sale of any security, financial instrument, financial product or service, or to be used in any way for evaluating the merits of participating in any transaction, and shall not constitute a solicitation under any jurisdiction or to any person, if such solicitation under such jurisdiction or to such person would be unlawful.

\begin{acks}
This research was funded in part by JPMorgan Chase Co.
\end{acks}

\bibliographystyle{abbrv}
\bibliography{mybib}

\end{document}

%% file: convergence.tex
  \begin{figure*}
\centering
  \begin{subfigure}{0.22\textwidth}
  \includegraphics[width=\textwidth]{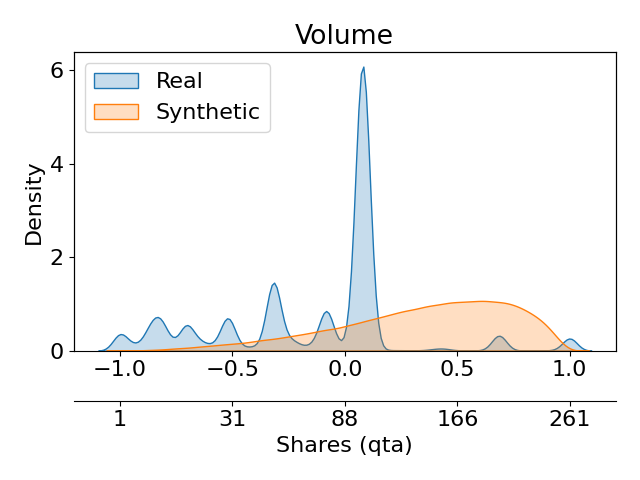}
  \end{subfigure}
  \hfill
\centering
  \begin{subfigure}{0.22\textwidth}
  \includegraphics[width=\textwidth]{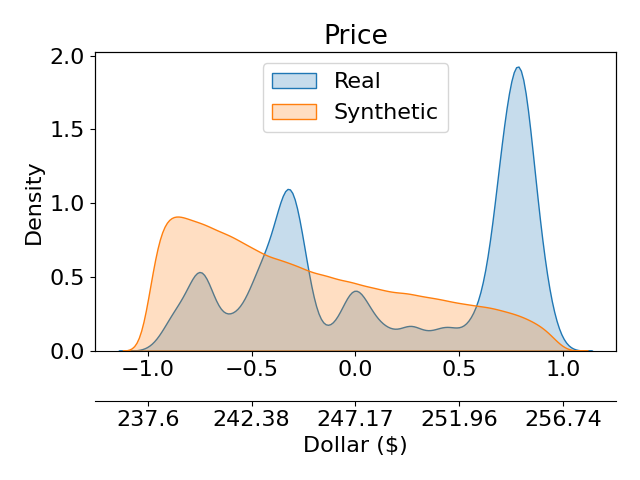}
  \end{subfigure} 
  \hfill
\centering
  \begin{subfigure}{0.22\textwidth} 
  \includegraphics[width=\textwidth]{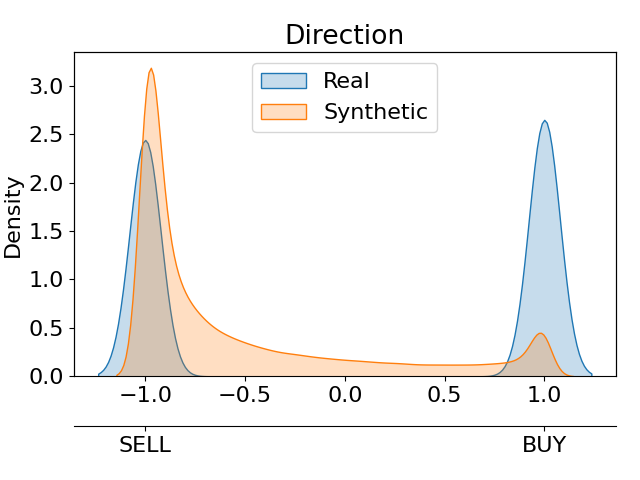} 
  \end{subfigure}  
  \hfill
\centering
  \begin{subfigure}{0.22\textwidth} 
  \includegraphics[width=\textwidth]{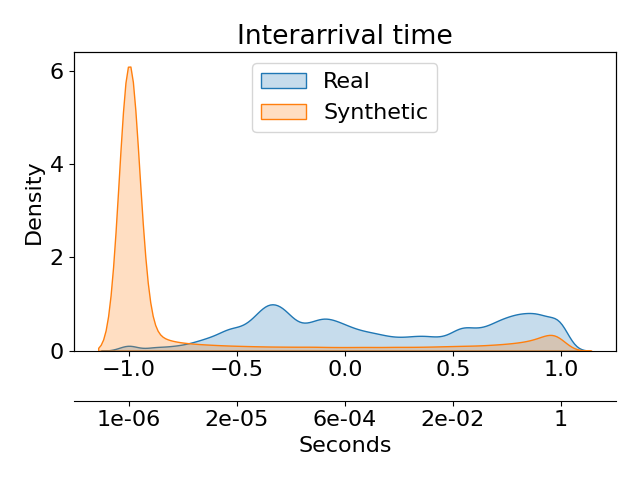} 
  \end{subfigure}
  
  \centering
 \begin{subfigure}{0.22\textwidth}
  \includegraphics[width=\textwidth]{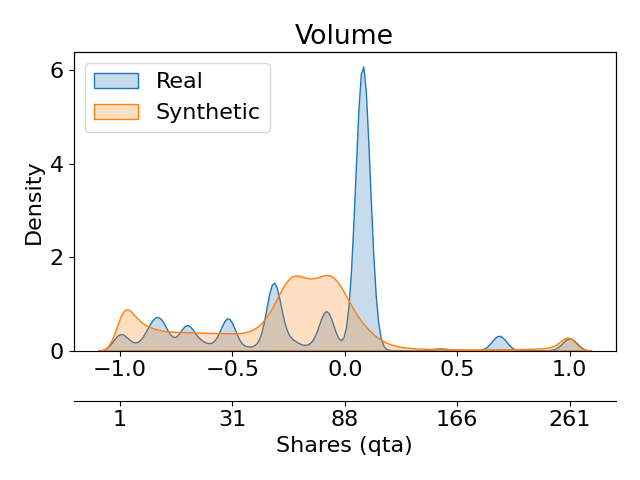}
  \end{subfigure}
  \hfill
 \centering
  \begin{subfigure}{0.22\textwidth}
  \includegraphics[width=\textwidth]{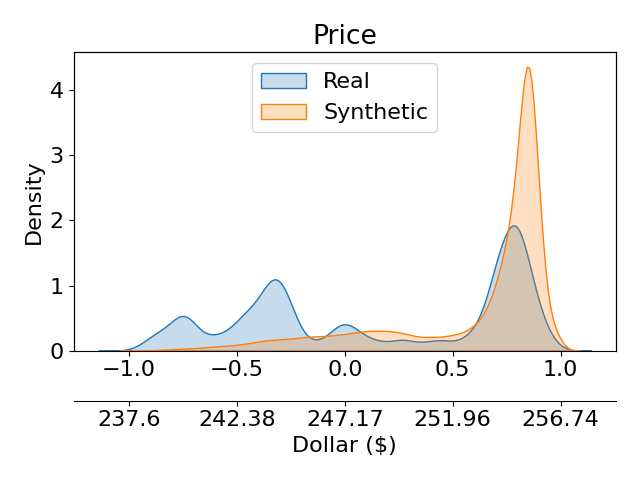}
  \end{subfigure} 
  \hfill
 \centering
  \begin{subfigure}{0.22\textwidth} 
  \includegraphics[width=\textwidth]{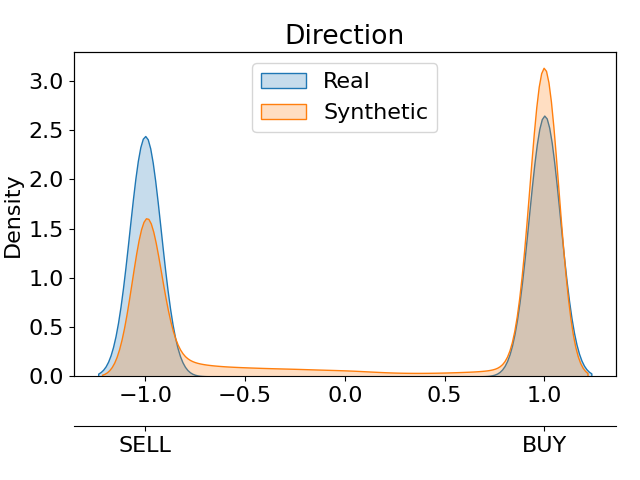} 
  \end{subfigure}  
 \hfill
 \centering
  \begin{subfigure}{0.22\textwidth} 
  \includegraphics[width=\textwidth]{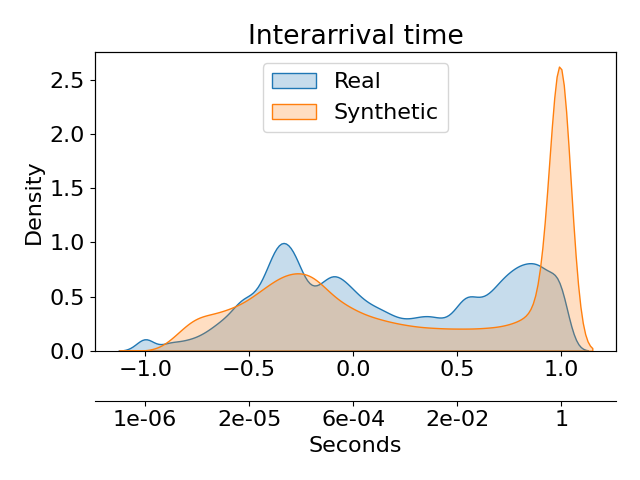} 
  \end{subfigure}
  \hfill
  
 \centering
  \begin{subfigure}{0.22\textwidth}
  \includegraphics[width=\textwidth]{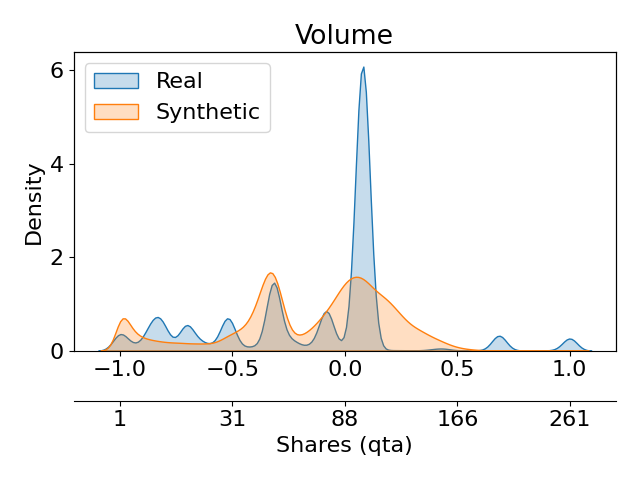}
  \end{subfigure}
 \hfill
 \centering
  \begin{subfigure}{0.22\textwidth}
  \includegraphics[width=\textwidth]{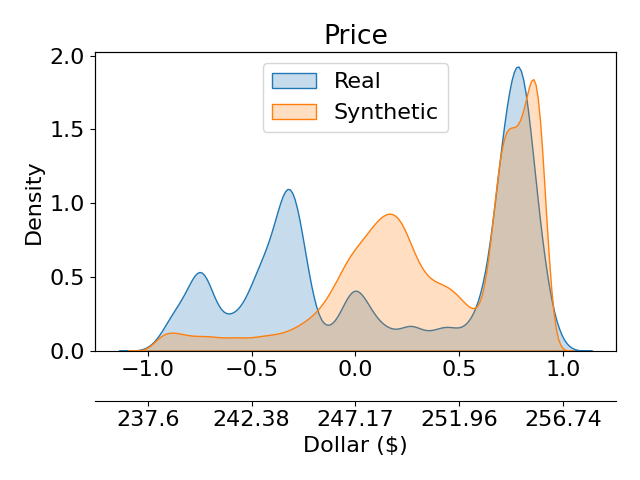}
  \end{subfigure} 
   \hfill
 \centering
  \begin{subfigure}{0.22\textwidth} 
  \includegraphics[width=\textwidth]{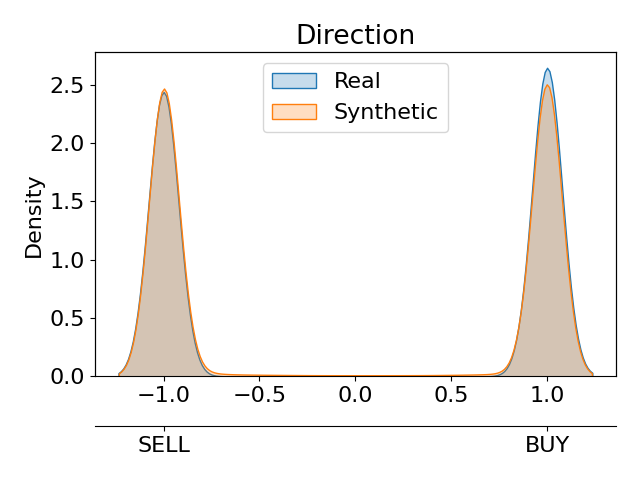} 
  \end{subfigure}  
 \hfill
 \centering
  \begin{subfigure}{0.22\textwidth} 
  \includegraphics[width=\textwidth]{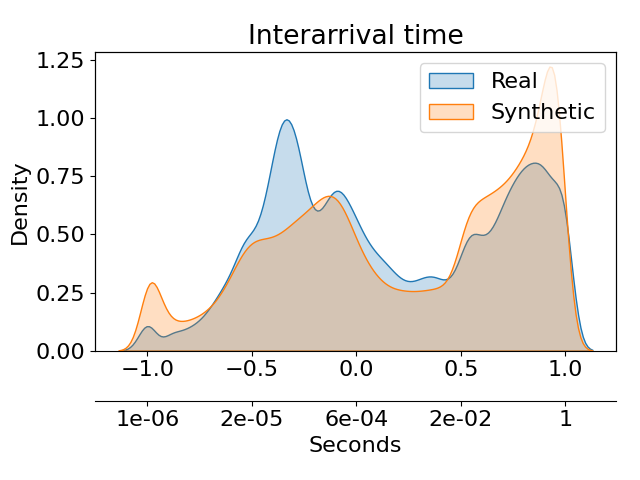} 
  \end{subfigure}
  
  \caption{Comparison of CGAN syntethic and real TSLA orders - Stylized Facts during the training process, at beginning (top), half (middle) and when it achieves reasonable convergence (bottom).}\label{fig:conv_study}
  \vspace{-0.05in}
  \end{figure*}
  

%% file: impact.tex
  \begin{figure}[t]
  \begin{subfigure}{0.49\columnwidth}
  \includegraphics[width=\textwidth]{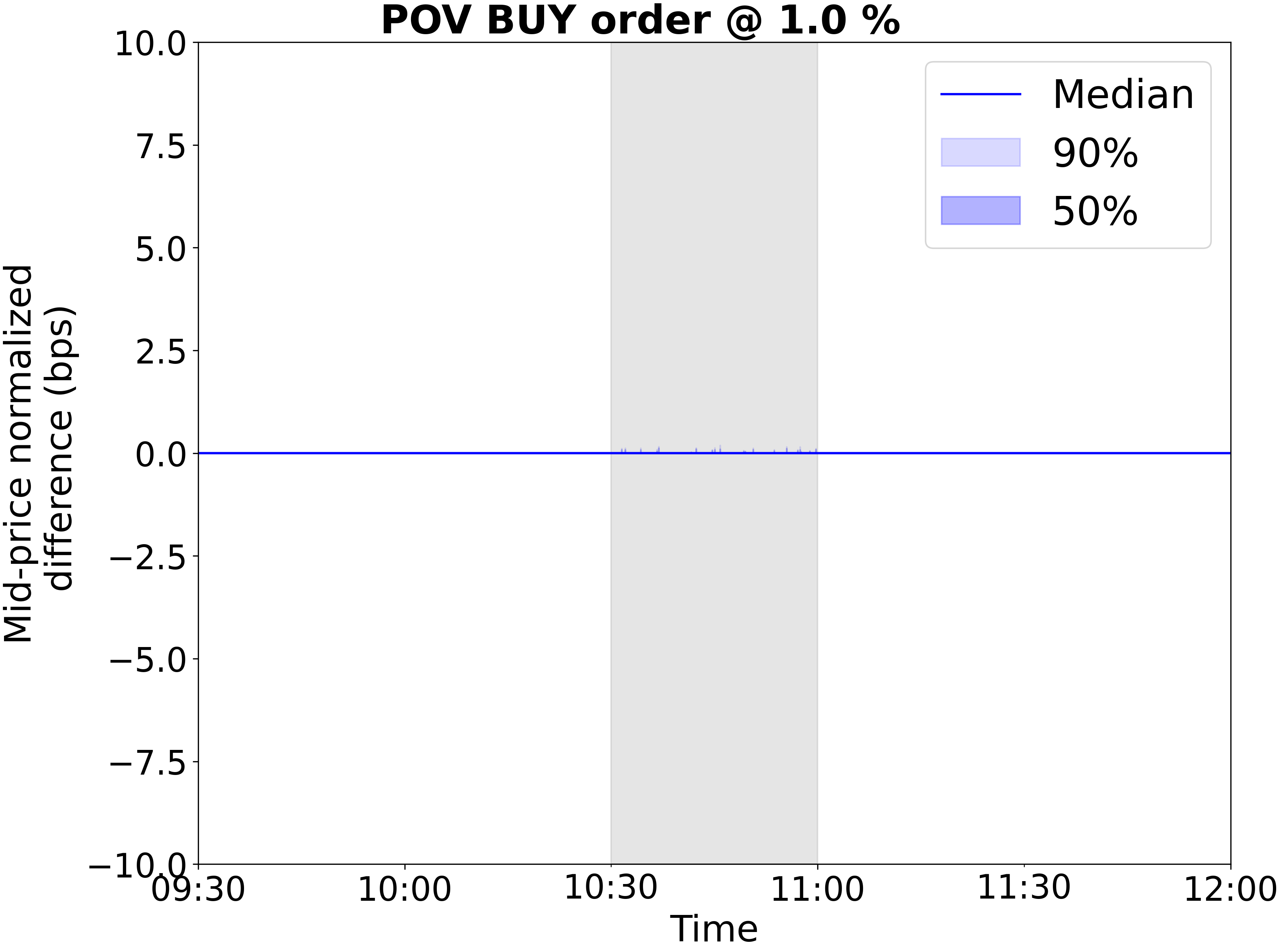}
  \end{subfigure}
  \hfill
  \begin{subfigure}{0.49\columnwidth}
  \includegraphics[width=\textwidth]{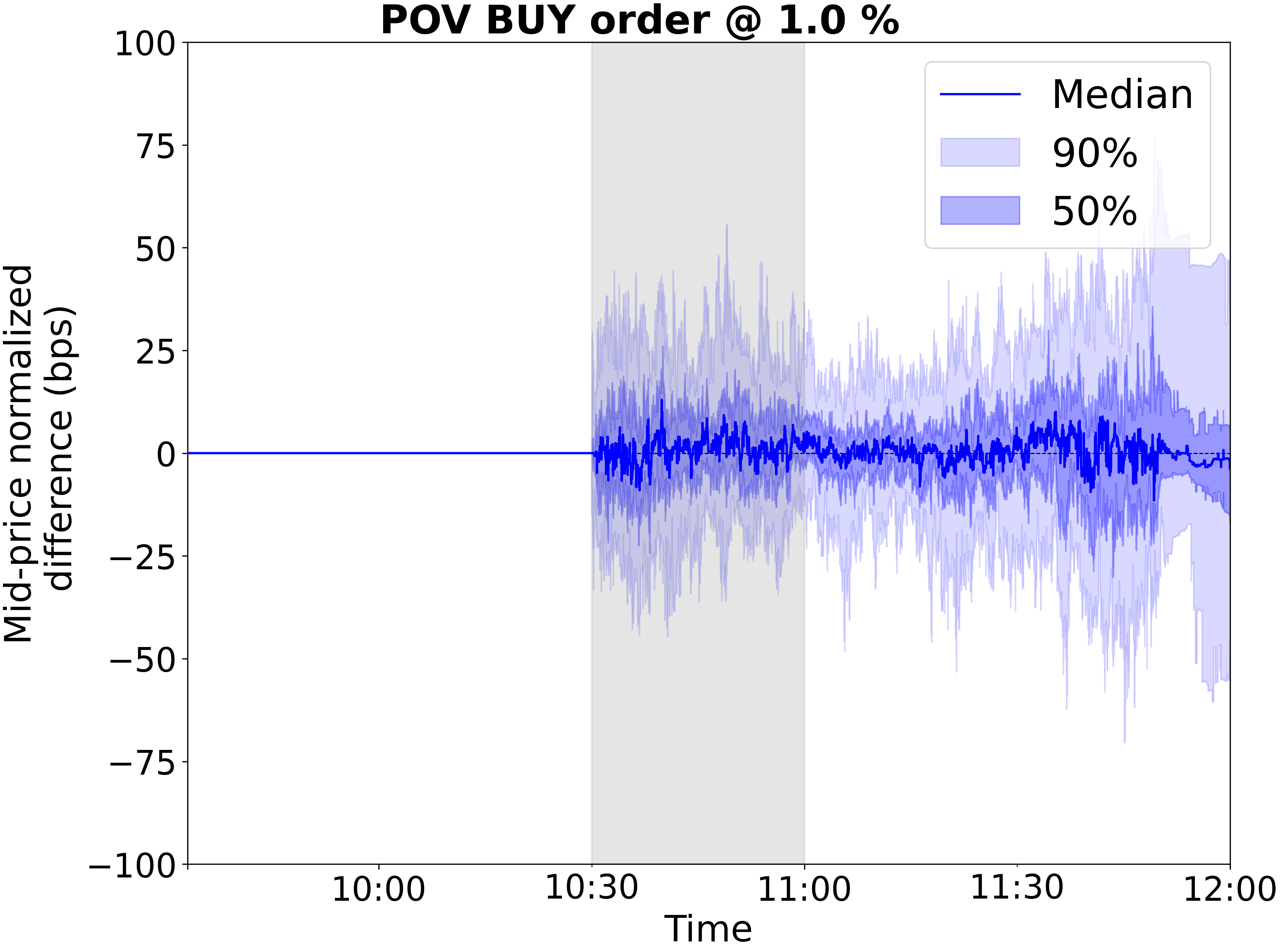}
  \end{subfigure} 
  \begin{subfigure}{0.49\columnwidth} 
  \includegraphics[width=\textwidth]{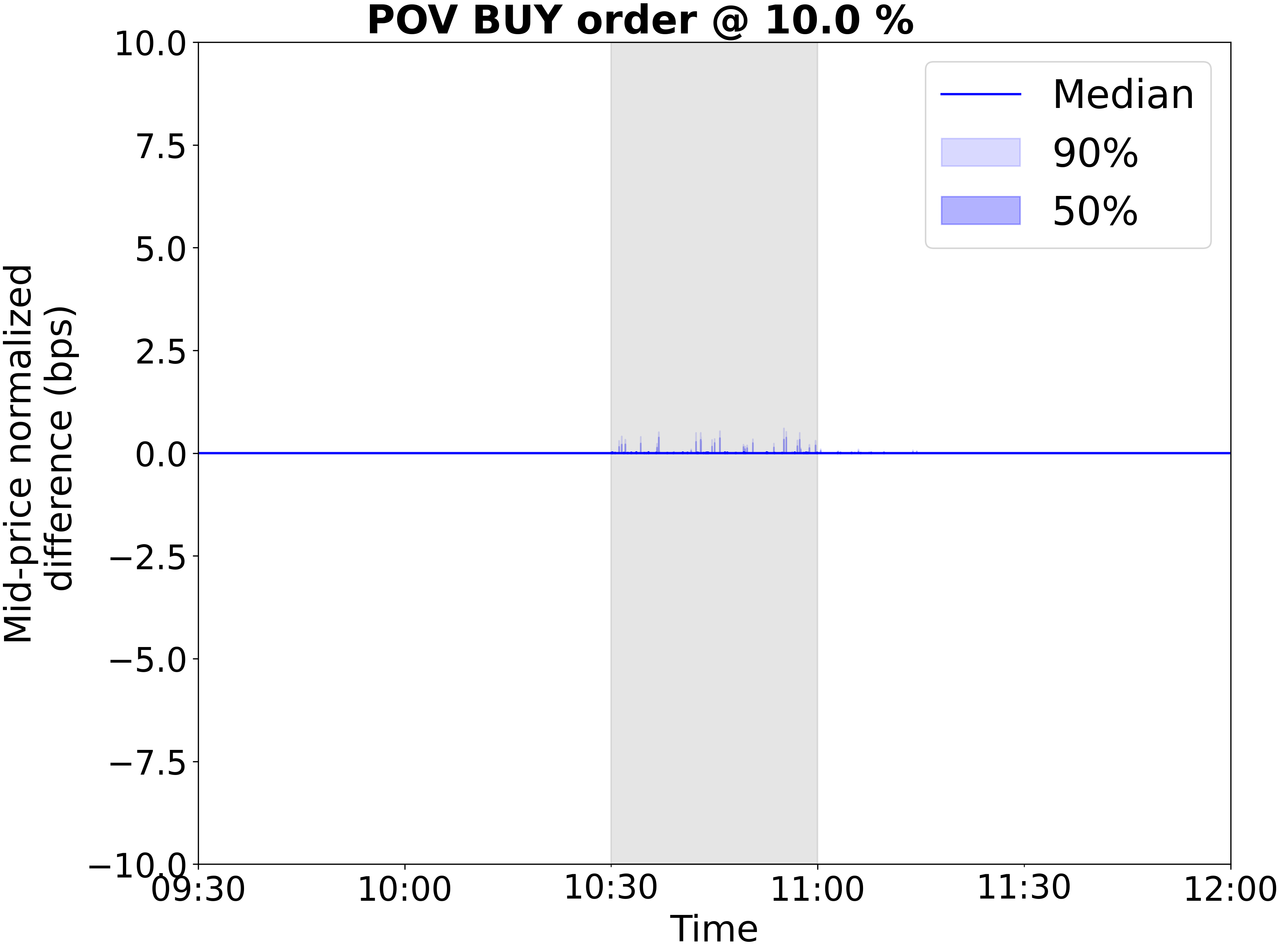} 
  \end{subfigure}  
  \hfill 
  \begin{subfigure}{0.49\columnwidth} 
  \includegraphics[width=\textwidth]{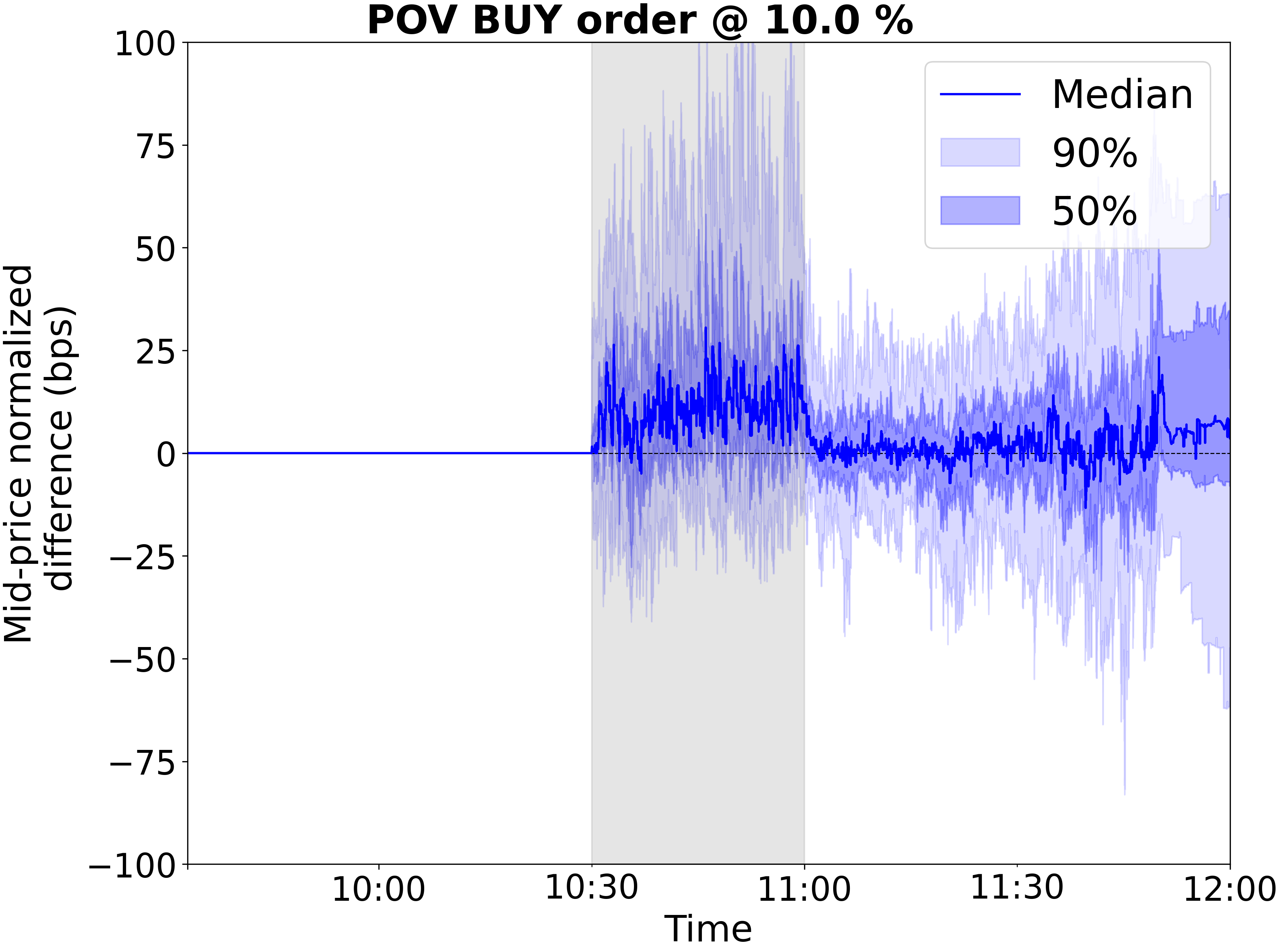} 
  \end{subfigure}
  
  \begin{subfigure}{0.49\columnwidth} 
  \includegraphics[width=\textwidth]{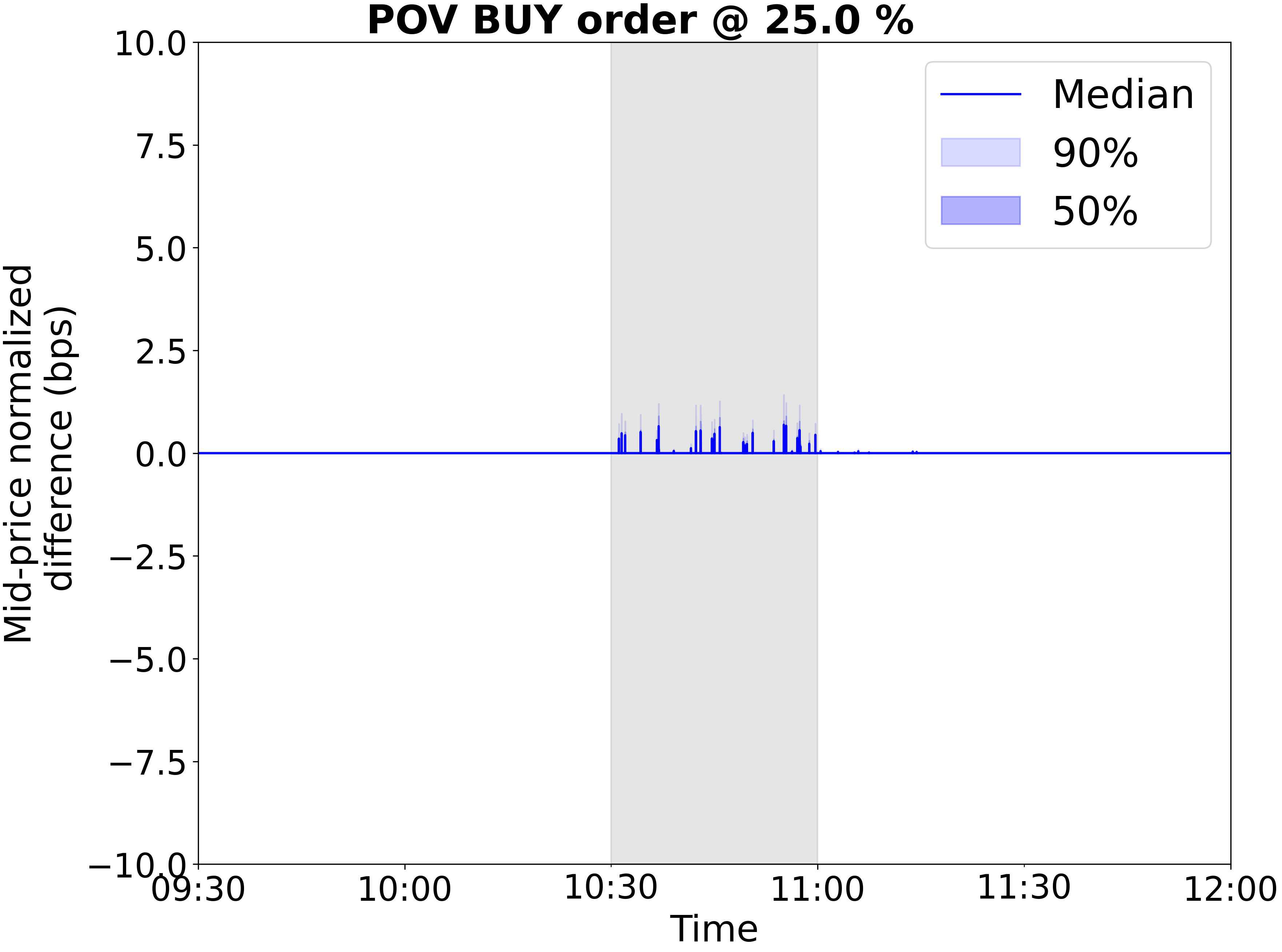} 
  \end{subfigure}  
  \hfill 
  \begin{subfigure}{0.49\columnwidth} 
  \includegraphics[width=\textwidth]{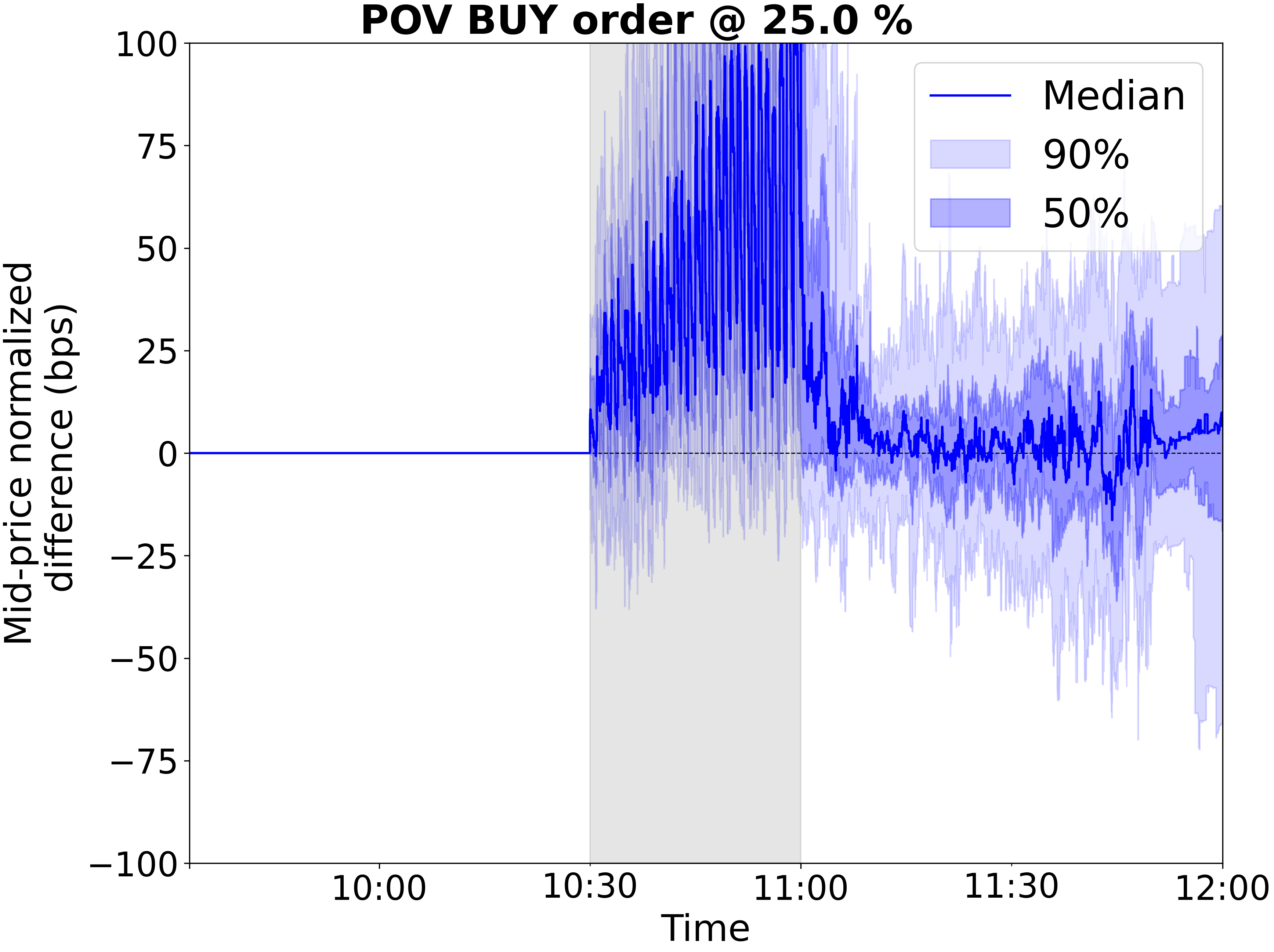} 
  \end{subfigure}
  
   \caption{Market replay (left) and CGAN-Based simulation (right) with and without a $\lambda$-POV execution agent. POV agent places orders between 10:30 AM and 11:00 AM with $\lambda \in [0.01, 0.1, 0.25]$, respectively on top, middle and bottom row. POV wake us frequency = 1 minute. The x-axes are truncated and y-axes expanded for clarity. } \label{fig:impact_plot}
   \vspace{-0.15in}
  \end{figure}

%% file: styl_facts.tex
  \begin{figure}
\centering
  \begin{subfigure}{0.22\textwidth} 
  \includegraphics[width=\textwidth]{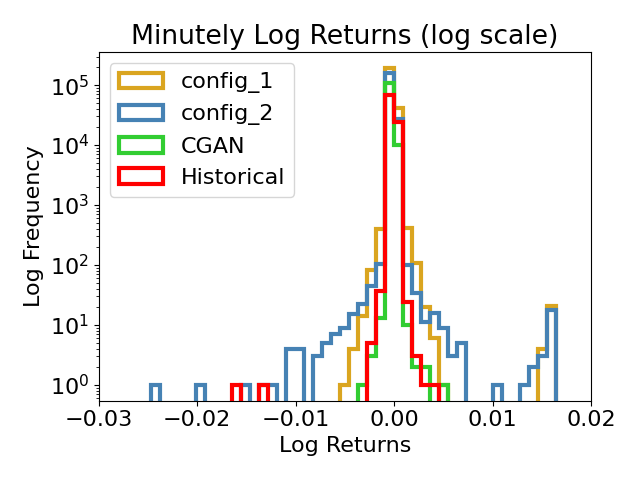} 
  \end{subfigure}
  \hfill
\centering
  \begin{subfigure}{0.22\textwidth}
  \includegraphics[width=\textwidth]{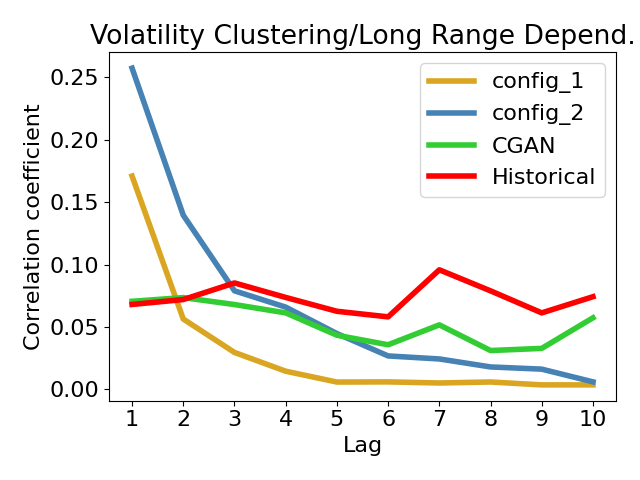}
  \end{subfigure} 
  \hfill
\centering
  \begin{subfigure}{0.22\textwidth} 
  \includegraphics[width=\textwidth]{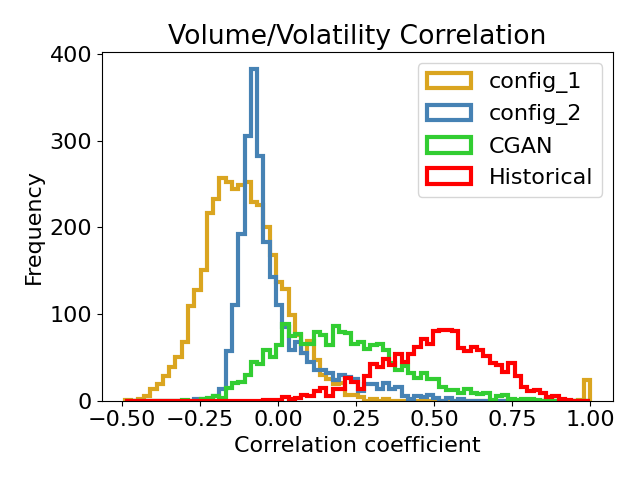} 
  \end{subfigure}  
  \hfill
  \centering
  \begin{subfigure}{0.22\textwidth}
  \includegraphics[width=\textwidth]{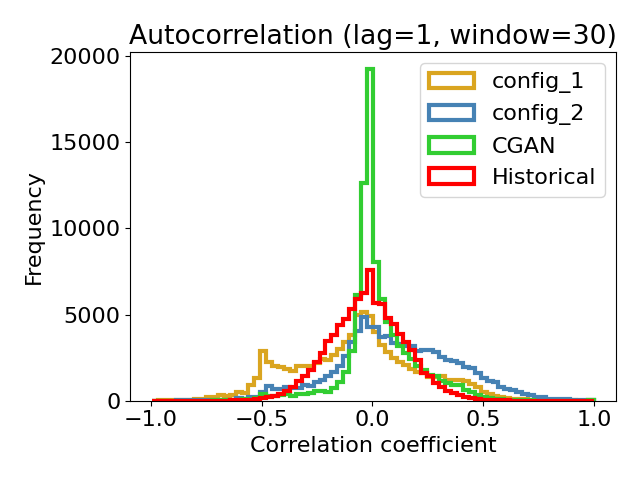}
  \end{subfigure}
  \vspace{-0.1in}
  \caption{Market Stylized facts (TSLA) - historical data against simulations.}\label{fig:styl_facts}
  \vspace{-0.15in}
  \end{figure}